\documentclass[preprint,12pt]{elsarticle}

\usepackage{amssymb}
\usepackage{pgfplots}
\usepackage{xcolor}
\usepackage{lineno}
\usepackage{amsmath, bm}
\usepackage{todonotes}
\usepackage{tabularx}

\begin{document}
\begin{frontmatter}

\title{Bayesian Optimization as a Flexible and Efficient Design Framework for Sustainable Process Systems}

\author[inst1]{Joel A. Paulson}

\affiliation[inst1]{organization={Department of Chemical and Biomolecular Engineering, The Ohio State University},
            addressline={151 W Woodruff Ave}, 
            city={Columbus, OH},
            postcode={43210}, 
            country={USA}}

\author[inst2]{Calvin Tsay\corref{cor}}

\affiliation[inst2]{organization={Department of Computing, Imperial College London},
            addressline={180 Queen's Gate}, 
            city={London, England},
            postcode={SW7 2AZ}, 
            country={UK}}
\cortext[cor]{Corresponding author: \texttt{c.tsay@imperial.ac.uk}}

\begin{abstract}
Bayesian optimization (BO) is a powerful technology for optimizing noisy expensive-to-evaluate black-box functions, with a broad range of real-world applications in science, engineering, economics, manufacturing, and beyond. In this paper, we provide an overview of recent developments, challenges, and opportunities in BO for design of next-generation process systems. After describing several motivating applications, we discuss how advanced BO methods have been developed to more efficiently tackle important problems in these applications. We conclude the paper with a summary of challenges and opportunities related to improving the quality of the probabilistic model, the choice of internal optimization procedure used to select the next sample point, and the exploitation of problem structure to improve sample efficiency. 
\end{abstract}


\end{frontmatter}


\section{Introduction}

Bayesian optimization (BO) refers to a class of algorithms built to efficiently find a global maximizer of a set of unknown functions $\bm{f}$
\begin{align} \label{eq:intro_bo}
    \bm{x}^\star \in \underset{\bm{x} \in \mathcal{X}} {\text{argmax}} ~ \bm{f}(\bm{x}),
\end{align}
where $\bm{x} \in \mathbb{R}^d$ is the set of $d$ design variables with feasible search space $\mathcal{X} \subseteq \mathbb{R}^d$, and $\bm{f} : \mathbb{R}^d \to \mathbb{R}^m$ is generally a vector-valued objective function with $m \geq 1$ outputs, each representing an individual performance metric. The search space $\mathcal{X}$ typically takes the form of simple box constraints $\mathcal{X} = \{ \bm{x} \in \mathbb{R}^d : \bm{x}_L \leq \bm{x} \leq \bm{x}_U \}$, and the objective function $\bm{f}$ is assumed to have three main characteristics: (i) evaluations of $\bm{f}$ are expensive in terms of time and/or other resources; (ii) only zeroth-order information is provided when $\bm{f}$ is evaluated (derivative information is not available); and (iii) the observations of $\bm{f}$ may be corrupted by unknown random noise. Most BO methods consider single-objective problems with $m = 1$, such that $\bm{x}^\star$ satisfies $f(\bm{x}^\star) \geq f(\bm{x})$ for all $\bm{x} \in \mathcal{X}$ for scalar function $f : \mathbb{R}^d \to \mathbb{R}$~\citep{frazier2018tutorial}. In the multi-objective optimization (MOO) setting, there often is not a single $\bm{x}^\star$ that simultaneously maximizes all $m > 1$ objectives (see Section \ref{sec:directions}). 

Optimization of expensive, noisy, black-box functions commonly occurs in designing sustainable process systems; we review some motivating applications in Section \ref{sec:applications} below. 
In principle, one can apply any type of derivative-free optimization (DFO) method~\citep{van2022data} to tackle such problems; however, these methods may require a large number of evaluations to converge. When evaluations of $\bm{f}$ are expensive, we desire an intelligent sample selection strategy that accounts for all available information to select future samples. The BO framework provides a systematic and versatile way to identify highly informative design candidates using minimal function evaluations. This article reviews recent advances in BO methods and highlights their relevance to design of next-generation sustainable energy and process systems. We also offer some perspectives on future research directions and associated challenges.

\section{Motivating Applications} 
\label{sec:applications}
A ``design framework'' can be construed at various scales of sustainable process engineering. Before describing the BO framework more formally in Section \ref{sec:howBOworks}, some relevant example applications are presented here.
\\

\textit{Materials design.} The design of new materials (e.g., catalysts, membranes, and molecules) represents a difficult, constrained problem with multiple objectives related to performance, safety, profitability, sustainability, etc. 
Engineers have long used optimization schemes to guide search procedures--a paradigm known as computer-aided molecular design (CAMD). 
BO represents a promising avenue for CAMD in resource-limited settings since the cost (time and money) of material synthesis and testing is often high. There is also often added complexity in the problem setting such as multiple information sources (e.g., low- and high-fidelity molecular theories)~\citep{fromer2023computer,wang2022bayesian}. 
\\

\textit{Reaction design.} Engineering new reactions for sustainable processes is another area involving expensive black-box functions subject to noise due to disturbances, measurement techniques, and starting materials. 
Chemists typically experiment with new reactions in a lab environment based on a combination of their experience, mechanistic understanding, empirical data and simple heuristics~\citep{shields2021bayesian}. In combination with laboratory automation (often referred to as ``self-driving labs''), BO is an exciting direction for reducing the time and resources associated with experimental search and accelerating chemical discovery across of wide-variety of applications~\citep{shields2021bayesian,guo2023bayesian}. 
\\

\textit{Process design.} Once a material and reaction chemistry have been selected, the design of the associated protocols, equipment, and processes is a critical next step complicated by having many interrelated design decisions and constraints, e.g., safety and regulatory requirements~\citep{tsay2018survey}. Computer simulation tools, including computational fluid dynamics (CFD) and process flowsheeting software (Aspen, gPROMS, etc.), are often used to construct detailed models here~\citep{hickman2022bayesian,savage2023multi}. 
When users cannot easily access the equation-oriented representations of the model, they must resort to simulation-based strategies such as BO to efficiently explore the design space. 
\\

\textit{Control system design.} Given a process, the selection and tuning of the associated control strategy is important for ensuring the process behaves as desired under major disturbances, equipment drift, and other forms of uncertainty. Traditionally, complex controller tuning has been tackled using heuristic methods, though BO can substantially improve upon this, even in high-dimensional problems~\citep{paulson2023tutorial}. Due to its flexibility, BO can also be used to tackle realistic integrated design and control (IDC) problems~\citep{sorourifar2023computationally}, which have been identified as a grand challenge problem in process systems engineering~\cite{burnak2019towards}. 

\newcommand{\tabitem}{~~\llap{\textbullet}~~}
\newcolumntype{L}[1]{>{\raggedright\arraybackslash}p{#1}}

\section{How Does Bayesian Optimization Work?}
\label{sec:howBOworks}

BO views \eqref{eq:intro_bo} as a ``learning problem'' with the goal of gaining information about solutions of \eqref{eq:intro_bo} by collecting observations of $\bm{f}$. Let $\mathcal{D}_k = \{ (\bm{x}_i, \bm{y}_i) \}_{i=1}^k$ be a set of $k$ function evaluations $\bm{y}_i = \bm{f}( \bm{x}_i ) + \varepsilon_i$ where $\varepsilon_i$ is some random noise term. 
If the acquisition of data $\mathcal{D}_{k+1} = \mathcal{D}_k \cup \{ (\bm{x}_{k+1}, \bm{y}_{k+1}) \}$ is a Markov process, then the optimal policy $\bm{\pi}^\star: \mathcal{D}_k \rightarrow \bm{x}_{k+1}$ must satisfy a Bellman equation derived from the principle of optimality~\citep{lam2016bayesian}. 
The well-known curse of dimensionality prevents us from solving the Bellman equation exactly, so we must resort to (hopefully useful) approximations. We argue popular BO methods satisfy this criteria; they are approximations that effectively navigate the tradeoff between exploration (learning for the future) and exploitation (immediately advancing toward the goal). 

\begin{figure}[ht!]
\centering
    \includegraphics[width=\textwidth]{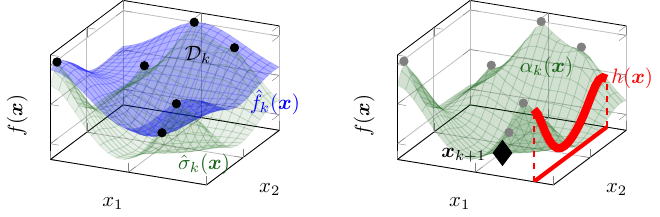}
    \vspace{-24pt}
    \caption{Two key steps of Bayesian optimization for minimizing a function (i.e., maximising its negation). Left: regressing a surrogate model with uncertainty. Right: optimizing an acquisition function (subject to optional constraints).}
    \label{fig:bo-illustration}
\end{figure}

BO methods have two main components, shown in Figure~\ref{fig:bo-illustration}: (i) a predictive surrogate model equipped with uncertainty estimates  (typically described by Bayesian statistics) and (ii) a so-called \textit{acquisition function}, or AF, whose value at any query point $\bm{x} \in \mathcal{X}$ quantifies the ``value'' of evaluating $\bm{f}$ at this point given existing observations $\mathcal{D}_k$. 
Let $\hat{\bm{f}}_k(\bm{x}) \in \mathbb{R}^m$ and $\hat{\Sigma}_k(\bm{x}) \in \mathbb{S}_+^m$ denote the surrogate mean and covariance predictors for $\bm{f}(\bm{x})$ given data $\mathcal{D}_k$. These quantities can be used to construct an uncertainty set $\mathbb{F}_k(\bm{x})$. For example, for a multivariate normal distribution, one can select an $\alpha$-level confidence region:
\begin{align} \label{eq:confidence-region}
    \mathbb{F}_k(\bm{x}) = \{ \bm{y} \in \mathbb{R}^m : ( \bm{y} - \hat{\bm{f}}_k(\bm{x}) )^\top \hat{\Sigma}^{-1}_k(\bm{x}) ( \bm{y} - \hat{\bm{f}}_k(\bm{x}) ) \leq r(\alpha) \},
\end{align}
where $r(\alpha)$ is the radius of the ellipsoid that corresponds to the cumulative $\chi^2$ distribution with $m$ degrees of freedom evaluated at confidence level $\alpha \in (0,1)$. The choice of predictive model is a very important component of BO, as performance relies on the quality of the model used to select sample locations. 
Gaussian processes (GPs) are by far the most commonly used surrogate model in BO due to their ability to analytically propagate uncertainty and their non-parametric nature. 
Confidence bounds generated by the GP can also be shown to rigorously bound the true unknown function as long as it belongs to a certain reproducing kernel Hilbert space (RKHS)~\citep{fiedler2021practical}, implying $\bm{f}(\bm{x}) \in \mathbb{F}_k(\bm{x})$ for a sufficiently large $\alpha$. 

Given a surrogate model $\hat{\bm{f}}_k(\bm{x})$ and $\hat{\Sigma}_k(\bm{x})$, we must then select a particular AF $\alpha_k : \mathcal{X} \to \mathbb{R}$, where the subscript $k$ highlights its dependence on $\mathcal{D}_k$. The BO sampling policy can then be formally written as
\begin{align} \label{eq:optimization-acq}
    \bm{x}_{k+1} \in \text{argmax}_{\bm{x} \in \mathcal{X}}~\alpha_k( \bm{x} ).
\end{align}
One can actually map the choice of AF to that of a reward function $R( \mathcal{D}_k )$. Specifically, we can think of $\alpha_k$ as the one-step expected increase in reward:
\begin{align} \label{eq:acquisition-func}
    \alpha_k(\bm{x}) = \mathbb{E}_k \left\lbrace R( \mathcal{D}_k \cup \{ \bm{x}, \bm{y}_{\bm{x}} \} ) - R( \mathcal{D}_k ) \right\rbrace,
\end{align}
where $\mathbb{E}_k$ refers to the expectation under the posterior distribution given $\mathcal{D}_k$. For example, the popular AFs expected improvement (EI) and knowledge gradient (KG) can be derived for the single-objective case by setting the reward function equal to the best observed sample $R( \mathcal{D}_k ) = \max_{(x,y)\in \mathcal{D}_k} y$ and maximum posterior mean value $R(\mathcal{D}_k) = \max_{x \in \mathcal{X}} \mathbb{E}\{ f(x) \mid \mathcal{D}_k \}$, respectively.

Following the discussion above, BO represents an optimal data acquisition strategy under the following four assumptions:
\begin{enumerate}
    \item The probabilistic surrogate model is optimally calibrated;
    \item The acquisition function $\alpha_k$ matches the desired reward function $R(\mathcal{D}_k)$; 
    \item The global maximizer of the acquisition function in \eqref{eq:optimization-acq} is found; 
    \item Only one step remains in the function evaluation budget, i.e., $N=1$. 
\end{enumerate}
It remains an open question as to why (and under what conditions) BO works well in practice, even when one or more of these assumptions are not satisfied. Along the same lines, it is not known how much is lost in performance due to approximations made in each of these areas. 
However, it has become clear in recent years that significant gains may be achieved by revisiting the assumptions of standard BO to better align with real-world problems. We review some of these emerging directions in the next section. 

\section{Emerging Directions in Bayesian Optimization}
\label{sec:directions}

This section reviews recent research directions related to BO that have particular relevance to the design of sustainable process systems. 
We start with some common problem characteristics (high dimensionality, discrete decisions, multiple objectives), then discuss constraints that arise, and conclude with some experimental considerations (batch and multi-fidelity experiments). 
Table \ref{tab:challenges} summarizes how these challenges affect the two BO steps described in Section \ref{sec:howBOworks}. 
Note that the above applications can present more than one of the below challenges, and likewise many of the reviewed works address multiple challenges simultaneously. 
We aim to provide a brief survey of recent research; interested readers are referred to~\citep{wang2023recent} for a more comprehensive survey and taxonomy of BO algorithms.
\\

\begin{table}[!h]
\caption{Summary of technical difficulties in addressing new BO problem features.}
\label{tab:challenges}
\vspace{8pt}
\scriptsize
    \begin{tabular}{L{3cm}|L{4.5cm}|L{4.5cm}}
        Problem feature & Surrogate model & Acquisition function \\ \hline
        High-dimensional ($d > 20$) & Poor sample coverage makes it hard to avoid overfitting & Difficult to optimize AF over many dimensions at same time \\ \hline
        Hybrid search space (e.g., $x_i \in \{0,1\}$) & Must build model over non-smooth hybrid input space & AF optimization is mixed-integer nonlinear program (MINLP) \\ \hline
        Multi-objective ($m > 1$) & Must build multi-output model with unknown correlations & AF optimization requires characterization of Pareto frontier \\ \hline
        Physics knowledge (e.g., $\bm{f}(\bm{x}) = \bm{w}(\bm{b}(\bm{x}))$) & Same as multi-objective since inner multi-output $\bm{b}$ is unknown & AF optimization challenging due to non-Gaussian predictions of $\bm{f}$ \\ \hline
        Black-box constraints (e.g., $\bm{g}(\bm{x}) \leq \bm{0}$) & Must learn safe/unsafe regions in addition to objectives & Constrained AF optimization to tradeoff performance/feasibility \\ \hline
        Path constraints ($\bm{x}_{k+1} \in \mathcal{T}_k(\bm{x}_k)$) & May potentially require both local and global surrogates & Must deal with lookahead AF optimization problem \\ \hline
        Multi-fidelity ($[ f_1(\bm{x}), ..., f_m(\bm{x}) ]$) & Need to decide how many sources and their degree of correlation & Joint AF optimization over both design and fidelity parameters \\ \hline
        Batch setting ($\bm{X}_t = [ \bm{x}_{t}^1,..., \bm{x}_{t}^b ]$) &Modeling task is the same as traditional BO & AF optimization over large set of parallel design parameters \\
    \end{tabular}
\end{table}

\textit{High-dimensional search spaces.} Many design problems, especially those that consider multiple space- and time-scales, involve the simultaneous selection of a large number of input variables, i.e., the dimensionality of $\bm{x}$ in \eqref{eq:intro_bo} is large $d \gg 1$. 
For instance, design of a chemical process may involve tens--or even hundreds--of decision variables related to process flows, equipment specifications, and operating conditions. 
However, in practice, many BO methods are limited to a relatively small number of dimensions $d \sim 10$~\citep{moriconi2020high}, and the ML community has dedicated many efforts to address scalability of BO (see \citet{malu2021bayesian} for a recent review). 
One promising direction here is to perform BO over a lower-dimensional, latent subspace~\citep{moriconi2020high,eriksson2021high,letham2020re}, in a similar spirit to projection-based methods for black-box optimization~\citep{bajaj2019unipopt}. 
These methods may exploit known structure or governing laws of an underlying \textit{physical} system when creating such subspaces. 
An alternative direction is to employ trust-region methods~\citep{eriksson2019scalable}, where multiple local models are employed to model ``trust regions'' of the full search space. 
\\

\textit{Hybrid discrete search spaces.} Discrete or categorical input variables can arise when selecting from alternative materials, a number of equipment, etc. (sometimes termed process `synthesis'). For instance, the choice of \textit{which} atom to place at a position rather than \textit{how much}. In combination with standard continuous inputs, such as temperatures and pressures, this presents a \textit{hybrid} (discrete/continuous) search space. 
This problem is often addressed by engineering new GP kernels that can handle discrete inputs, such as diffusion kernels~\citep{deshwal2021bayesian}, tree-based kernels~\citep{thebelt2022tree}, or hybrid discrete-continuous kernels~\citep{ru2020bayesian}. 
An alternative is to use non-GP surrogate models, such as parametric models~\citep{daxberger2021mixed} or tree ensembles~\citep{thebelt2022multi}. 
\citet{dreczkowski2023framework} provide a set of benchmarks for discrete and hybrid BO. 
Future progress is still required to make these methods more scalable, e.g., to high-dimensional problems or constrained settings. 
\\

\textit{Multiple objectives.} Designing sustainable processes, materials, and/or chemistries often involves multiple competing objectives. For exmample, engineering a new product material may require balancing sustainability, profitability, and robustness objectives~\citep{fromer2023computer,rangaiah2020multi}. 
In these cases, the objective becomes to find a set of optimal points that are Pareto optimal solutions, defined as points where no objective can be improved without worsening another~\citep{wang2023recent}. 
The process systems literature adjusts BO to produce Pareto points via Tchebyshev scalarizations~\citep{thebelt2022multi}, multi-objective Thompson sampling~\citep{folch2023practical}, or $\epsilon$-constraint approaches~\citep{beykal2018optimal}. 
Alternatively, BO can be modified to employ surrogates that directly model multiple objectives, enabling the use of new AFs such as probability of hypervolume improvement or expected hypervolume improvement~\citep{fromer2023pareto}. 
\\

\textit{White-box constraints and physics-based knowledge.} 
In many engineering applications, additional inequality $\bm{g}(\bm{x}) \leq \bm{0}$ and/or equality $\bm{h}(\bm{x})=\bm{0}$ constraints may limit the input search space $\mathcal{X}$. When these constraints are explicitly known (i.e., white-box functions), they can be directly incorporated into the AF optimization step~\citep{thebelt2022tree}. However, in the more general case, only partial information may be available for the constraints (and/or the objective). This so-called grey-box problem setting can be tackled by assuming some structure, e.g., composite functions $\bm{g}(\bm{x}) = \bm{w}(\bm{b}(\bm{x}))$ where $\bm{w}(\cdot)$ is the known (white-box) outer function and $\bm{b}(\cdot)$ is the unknown (black-box) inner function. 
For example, the former can describe known conservation laws that must be satisfied while the latter represent the unknown simulator or experiment that provides its input. Constrained BO with composite functions has been studied in~\citep{paulson2022cobalt}. One can then further generalize the description of $\bm{w}$ to be a network (or graph) of functions~\citep{astudillo2021bayesian, gonzalez2023bois}.
\\

\textit{Black-box constraints.}  
In practical settings, constraints may be fully unknown, similar to the objective functions $\bm{f}$. For example, when designing a new drug, toxicity is difficult to predict \textit{a priori}, so it is commonly estimated through a time-consuming synthesis and testing procedure. 
Research in this area can be roughly divided into two categories. 
The first category involves ``safe BO'' methods that attempt to limit constraint violations during optimization, such as to prescribe safety/feasibility in real-world experiments. 
For example, SafeOpt~\citep{berkenkamp2023bayesian} addresses this problem by combining GP confidence bounds \eqref{eq:confidence-region} with an explicit characterization of the set of safe points. 
Later research introduces more tractable approximations to the safety region~\citep{krishnamoorthy2022safe}, and related optimization strategies~\citep{chanCDCsafeBO}. 
The second category of methods does not require constraints to be satisfied at every iteration, which is more applicable for, e.g., computer-aided experiments. Here, there is a further distinction: `implicit' constrained BO methods encode the constraint using some merit function, such as an augmented Lagrangian~\citep{picheny2016bayesian} or the probability of feasibility~\citep{gardner2014bayesian}, while `explicit' constrained BO methods instead directly restrict the feasible region when solving \eqref{eq:optimization-ace}, such as by using the mean prediction of the constraint function or a relaxation thereof~\citep{lu2022no, LU2023103085}. 
\\

\textit{Path constraints.} 
When designing a sequence of experiments, it can take a significant amount of time (or create unsafe behaviors) when moving between two arbitrary design conditions, resulting in a restriction based on the previous experiment: $\bm{x}_{k+1} \in \mathcal{T}_k( \bm{x}_k )$. These path constraints can be explicitly accounted for using (approximate) dynamic-programming approaches \citep{paulson2022efficient, paulson2023lsr}, which can be further extended to handle black-box constraints~\citep{paulson2024self}. 
When path constraints arise from a \textit{movement cost}, we may handle this by (implicitly) limiting sequential experiments to be similar in a norm-sense, i.e., $\mathcal{T}_k( \bm{x}_k ) = \{ \bm{x}_{k+1} : \| \bm{x}_{k+1} - \bm{x}_k \| \leq b_k \}$. 
One such case is in designing a steady-state process; the system has to settle to a new steady state before taking a measurement. 
SnAKe~\citep{folch2022snake} and MONGOOSE~\citep{yang2023mongoose} address this problem implicitly by, respectively, ordering candidate experiments with an embedded traveling salesman problem and by meta-learning a parametric policy to generate smooth design trajectories. 
An alternative strategy is to directly minimize movement costs, which is presented and applied to optimize a wind energy system  in~\citep{ramesh2022movement}. 
\\

\textit{Multiple models or fidelities.} 
When sampling $f(\bm{x})$ in \eqref{eq:intro_bo}, we may have multiple experiment types available, such as physical vs computer experiments~\citep{savage2023multi}. 
For example, when designing a new catalyst, its performance could be estimated using different test reactors, with different measurement techniques, and/or with different levels of molecular simulations. 
In terms of BO, this gives a vector of functions $[ f_1(\bm{x}), ..., f_m(\bm{x}) ]$ that can be queried, typically with a tradeoff between accuracy (fidelity), and some measure of cost. 
Having samples of varying fidelities can be managed using (i) co-Kriging, or assuming that all samples (regardless of fidelity) follow a multivariate normal distribution, (ii) including fidelity as an additional \textit{latent} input varaible~\citep{foumani2023multi}, or (iii) assuming some structure among the functions, such as an increasing maximum error~\citep{folch2023combining}. 
Once the surrogate model is built, the AF-maximization step must not only select sampling locations, but also the model/experiment fidelities to sample given their associated costs. 
\\

\textit{Parallel/batch experiments.}
Process simulations and experiments can often be performed in parallel, i.e., a batch of experiments. For example, high-throughput experimental set-ups can enable testing many reaction conditions simultaneously. Methods for designing experimental batches from an AF are often limited to batches of ten samples or fewer (larger batches may motivate alternative black-box optimization methods). 
In addition to modifying existing AFs~\citep{zhou2023comprehensive} (e.g., extending to improvement of the batch), generating a batch from the AF can be performed by Thompson sampling~\citep{folch2023combining,hernandez2017parallel} or reformulating this as a Bayesian quadrature problem~\citep{adachi2023sober} (including hybrid input spaces and constraints~\citep{adachi2023domain}). 
The cost of each experiment can also be accounted for when designing a batch~\citep{liang2023capbo}. 
Future directions include faster or \textit{asynchronous} batch settings, where querying points may also involve different delays before observing the function output(s). 
\\

\section{Conclusions and Future Directions}
Bayesian optimization (BO) is a promising flexible framework for solving challenging design problems in sustainable process systems. These types of design problems can arise at or across a variety of time- and length-scales, from small-scale material design problems to large-scale process design problems.
As such, sustainable engineering applications present important and unique challenges such as high-dimensionality, mixed continuous-discrete decisions, multiple objectives, multiple information sources of varying costs, constraints, and batch (parallel) experimentation settings.
This paper reviews recent advances in BO that are relevant to sustainable process system design and beyond. We believe that the unique challenges of this area provide opportunities for new machine learning technology, as well as for engineers to contribute fundamental improvements to BO.

There are several exciting future research directions, which we broadly categorize into three areas: applications, methods, and theory. First, due to the recent explosion in novel BO methods, there would be significant value in their application to important and existing problems in sustainable engineering. 
Development of a systematic and fair performance comparison methodology (especially in practical applications with limited/noisy data) is an interesting problem that has seen relatively little attention. Second, there is a need for methods that can handle several challenges described in Section \ref{sec:directions} simultaneously (e.g., multi-objective, physics-based, constrained, and multi-fidelity), as many real-world problems naturally involve a combination of them. There are also new problem settings that have received relatively little attention such as BO with human preferences, BO over distributed graph networks, causal BO, and meta-learning for BO. Lastly, as mentioned in Section \ref{sec:howBOworks}, the vast majority of BO methods are suboptimal compared to the dynamic programming solution. New theory that can establish suboptimality bounds---even for specific cases---would be valuable to help guide the choice of approximate policies or to demonstrate to practitioners that some directions are fundamentally worse than others for certain classes of problems.

\section*{Acknowledgements}
The authors gratefully acknowledge funding from the National Science Foundation (award number 2237616) and the Engineering and Physical Sciences Research Council (fellowship award EP/T001577/1). The authors also thank Jose Pablo Folch for the insightful feedback and discussions.

\newpage
\bibliographystyle{elsarticle-num-names} 
\bibliography{borefs}

\begin{thebibliography}{54}
\expandafter\ifx\csname natexlab\endcsname\relax\def\natexlab#1{#1}\fi
\providecommand{\url}[1]{\texttt{#1}}
\providecommand{\href}[2]{#2}
\providecommand{\path}[1]{#1}
\providecommand{\DOIprefix}{doi:}
\providecommand{\ArXivprefix}{arXiv:}
\providecommand{\URLprefix}{URL: }
\providecommand{\Pubmedprefix}{pmid:}
\providecommand{\doi}[1]{\href{http://dx.doi.org/#1}{\path{#1}}}
\providecommand{\Pubmed}[1]{\href{pmid:#1}{\path{#1}}}
\providecommand{\bibinfo}[2]{#2}
\ifx\xfnm\relax \def\xfnm[#1]{\unskip,\space#1}\fi
\bibitem[{Frazier(2018)}]{frazier2018tutorial}
\bibinfo{author}{P.~I. Frazier},
\newblock \bibinfo{title}{A tutorial on {B}ayesian optimization},
\newblock \bibinfo{journal}{arXiv preprint arXiv:1807.02811}
  (\bibinfo{year}{2018}).
\bibitem[{van~de Berg et~al.(2022)van~de Berg, Savage, Petsagkourakis, Zhang,
  Shah, and del Rio-Chanona}]{van2022data}
\bibinfo{author}{D.~van~de Berg}, \bibinfo{author}{T.~Savage},
  \bibinfo{author}{P.~Petsagkourakis}, \bibinfo{author}{D.~Zhang},
  \bibinfo{author}{N.~Shah}, \bibinfo{author}{E.~A. del Rio-Chanona},
\newblock \bibinfo{title}{Data-driven optimization for process systems
  engineering applications},
\newblock \bibinfo{journal}{Chemical Engineering Science} \bibinfo{volume}{248}
  (\bibinfo{year}{2022}) \bibinfo{pages}{117135}.
\bibitem[{Fromer and Coley(2023)}]{fromer2023computer}
\bibinfo{author}{J.~C. Fromer}, \bibinfo{author}{C.~W. Coley},
\newblock \bibinfo{title}{Computer-aided multi-objective optimization in small
  molecule discovery},
\newblock \bibinfo{journal}{Patterns} \bibinfo{volume}{4}
  (\bibinfo{year}{2023}).
\bibitem[{Wang and Dowling(2022)}]{wang2022bayesian}
\bibinfo{author}{K.~Wang}, \bibinfo{author}{A.~W. Dowling},
\newblock \bibinfo{title}{Bayesian optimization for chemical products and
  functional materials},
\newblock \bibinfo{journal}{Current Opinion in Chemical Engineering}
  \bibinfo{volume}{36} (\bibinfo{year}{2022}) \bibinfo{pages}{100728}.
\bibitem[{Shields et~al.(2021)Shields, Stevens, Li, Parasram, Damani, Alvarado,
  Janey, Adams, and Doyle}]{shields2021bayesian}
\bibinfo{author}{B.~J. Shields}, \bibinfo{author}{J.~Stevens},
  \bibinfo{author}{J.~Li}, \bibinfo{author}{M.~Parasram},
  \bibinfo{author}{F.~Damani}, \bibinfo{author}{J.~I.~M. Alvarado},
  \bibinfo{author}{J.~M. Janey}, \bibinfo{author}{R.~P. Adams},
  \bibinfo{author}{A.~G. Doyle},
\newblock \bibinfo{title}{Bayesian reaction optimization as a tool for chemical
  synthesis},
\newblock \bibinfo{journal}{Nature} \bibinfo{volume}{590}
  (\bibinfo{year}{2021}) \bibinfo{pages}{89--96}.
\bibitem[{Guo et~al.(2023)Guo, Rankovi{\'c}, and Schwaller}]{guo2023bayesian}
\bibinfo{author}{J.~Guo}, \bibinfo{author}{B.~Rankovi{\'c}},
  \bibinfo{author}{P.~Schwaller},
\newblock \bibinfo{title}{Bayesian optimization for chemical reactions},
\newblock \bibinfo{journal}{Chimia} \bibinfo{volume}{77} (\bibinfo{year}{2023})
  \bibinfo{pages}{31--31}.
\bibitem[{Tsay et~al.(2018)Tsay, Pattison, Piana, and Baldea}]{tsay2018survey}
\bibinfo{author}{C.~Tsay}, \bibinfo{author}{R.~C. Pattison},
  \bibinfo{author}{M.~R. Piana}, \bibinfo{author}{M.~Baldea},
\newblock \bibinfo{title}{A survey of optimal process design capabilities and
  practices in the chemical and petrochemical industries},
\newblock \bibinfo{journal}{Computers \& Chemical Engineering}
  \bibinfo{volume}{112} (\bibinfo{year}{2018}) \bibinfo{pages}{180--189}.
\bibitem[{Hickman et~al.(2022)Hickman, Aldeghi, H{\"a}se, and
  Aspuru-Guzik}]{hickman2022bayesian}
\bibinfo{author}{R.~J. Hickman}, \bibinfo{author}{M.~Aldeghi},
  \bibinfo{author}{F.~H{\"a}se}, \bibinfo{author}{A.~Aspuru-Guzik},
\newblock \bibinfo{title}{Bayesian optimization with known experimental and
  design constraints for chemistry applications},
\newblock \bibinfo{journal}{Digital Discovery} \bibinfo{volume}{1}
  (\bibinfo{year}{2022}) \bibinfo{pages}{732--744}.
\bibitem[{Savage et~al.(2023)Savage, Basha, McDonough, Matar, and del
  Rio~Chanona}]{savage2023multi}
\bibinfo{author}{T.~Savage}, \bibinfo{author}{N.~Basha},
  \bibinfo{author}{J.~McDonough}, \bibinfo{author}{O.~K. Matar},
  \bibinfo{author}{E.~A. del Rio~Chanona},
\newblock \bibinfo{title}{Multi-fidelity data-driven design and analysis of
  reactor and tube simulations},
\newblock \bibinfo{journal}{Computers \& Chemical Engineering}
  (\bibinfo{year}{2023}) \bibinfo{pages}{108410}.
\bibitem[{Paulson et~al.(2023)Paulson, Sorourifar, and
  Mesbah}]{paulson2023tutorial}
\bibinfo{author}{J.~A. Paulson}, \bibinfo{author}{F.~Sorourifar},
  \bibinfo{author}{A.~Mesbah},
\newblock \bibinfo{title}{A tutorial on derivative-free policy learning methods
  for interpretable controller representations},
\newblock in: \bibinfo{booktitle}{Proceedings of the American Control
  Conference}, \bibinfo{organization}{IEEE}, \bibinfo{year}{2023}, pp.
  \bibinfo{pages}{1295--1306}.
\bibitem[{Sorourifar et~al.(2023)Sorourifar, Choksi, and
  Paulson}]{sorourifar2023computationally}
\bibinfo{author}{F.~Sorourifar}, \bibinfo{author}{N.~Choksi},
  \bibinfo{author}{J.~A. Paulson},
\newblock \bibinfo{title}{Computationally efficient integrated design and
  predictive control of flexible energy systems using multi-fidelity
  simulation-based {B}ayesian optimization},
\newblock \bibinfo{journal}{Optimal Control Applications and Methods}
  \bibinfo{volume}{44} (\bibinfo{year}{2023}) \bibinfo{pages}{549--576}.
\bibitem[{Burnak et~al.(2019)Burnak, Diangelakis, and
  Pistikopoulos}]{burnak2019towards}
\bibinfo{author}{B.~Burnak}, \bibinfo{author}{N.~A. Diangelakis},
  \bibinfo{author}{E.~N. Pistikopoulos},
\newblock \bibinfo{title}{Towards the grand unification of process design,
  scheduling, and control—utopia or reality?},
\newblock \bibinfo{journal}{Processes} \bibinfo{volume}{7}
  (\bibinfo{year}{2019}) \bibinfo{pages}{461}.
\bibitem[{Lam et~al.(2016)Lam, Willcox, and Wolpert}]{lam2016bayesian}
\bibinfo{author}{R.~Lam}, \bibinfo{author}{K.~Willcox}, \bibinfo{author}{D.~H.
  Wolpert},
\newblock \bibinfo{title}{Bayesian optimization with a finite budget: An
  approximate dynamic programming approach},
\newblock \bibinfo{journal}{Advances in Neural Information Processing Systems}
  \bibinfo{volume}{29} (\bibinfo{year}{2016}).
\bibitem[{Fiedler et~al.(2021)Fiedler, Scherer, and
  Trimpe}]{fiedler2021practical}
\bibinfo{author}{C.~Fiedler}, \bibinfo{author}{C.~W. Scherer},
  \bibinfo{author}{S.~Trimpe},
\newblock \bibinfo{title}{Practical and rigorous uncertainty bounds for
  {G}aussian process regression},
\newblock in: \bibinfo{booktitle}{Proceedings of the AAAI Conference on
  Artificial Intelligence}, volume~\bibinfo{volume}{35}, \bibinfo{year}{2021},
  pp. \bibinfo{pages}{7439--7447}.
\bibitem[{Wang et~al.(2023)Wang, Jin, Schmitt, and Olhofer}]{wang2023recent}
\bibinfo{author}{X.~Wang}, \bibinfo{author}{Y.~Jin},
  \bibinfo{author}{S.~Schmitt}, \bibinfo{author}{M.~Olhofer},
\newblock \bibinfo{title}{Recent advances in {Bayesian} optimization},
\newblock \bibinfo{journal}{ACM Computing Surveys} \bibinfo{volume}{55}
  (\bibinfo{year}{2023}) \bibinfo{pages}{1--36}.
\bibitem[{Moriconi et~al.(2020)Moriconi, Deisenroth, and
  Sesh~Kumar}]{moriconi2020high}
\bibinfo{author}{R.~Moriconi}, \bibinfo{author}{M.~P. Deisenroth},
  \bibinfo{author}{K.~Sesh~Kumar},
\newblock \bibinfo{title}{High-dimensional {B}ayesian optimization using
  low-dimensional feature spaces},
\newblock \bibinfo{journal}{Machine Learning} \bibinfo{volume}{109}
  (\bibinfo{year}{2020}) \bibinfo{pages}{1925--1943}.
\bibitem[{Malu et~al.(2021)Malu, Dasarathy, and Spanias}]{malu2021bayesian}
\bibinfo{author}{M.~Malu}, \bibinfo{author}{G.~Dasarathy},
  \bibinfo{author}{A.~Spanias},
\newblock \bibinfo{title}{Bayesian optimization in high-dimensional spaces: A
  brief survey},
\newblock in: \bibinfo{booktitle}{2021 12th International Conference on
  Information, Intelligence, Systems \& Applications (IISA)},
  \bibinfo{organization}{IEEE}, \bibinfo{year}{2021}, pp.
  \bibinfo{pages}{1--8}.
\bibitem[{Eriksson and Jankowiak(2021)}]{eriksson2021high}
\bibinfo{author}{D.~Eriksson}, \bibinfo{author}{M.~Jankowiak},
\newblock \bibinfo{title}{High-dimensional {B}ayesian optimization with sparse
  axis-aligned subspaces},
\newblock in: \bibinfo{booktitle}{Uncertainty in Artificial Intelligence},
  \bibinfo{organization}{PMLR}, \bibinfo{year}{2021}, pp.
  \bibinfo{pages}{493--503}.
\bibitem[{Letham et~al.(2020)Letham, Calandra, Rai, and Bakshy}]{letham2020re}
\bibinfo{author}{B.~Letham}, \bibinfo{author}{R.~Calandra},
  \bibinfo{author}{A.~Rai}, \bibinfo{author}{E.~Bakshy},
\newblock \bibinfo{title}{Re-examining linear embeddings for high-dimensional
  {Bayesian} optimization},
\newblock \bibinfo{journal}{Advances in Neural Information Processing Systems}
  \bibinfo{volume}{33} (\bibinfo{year}{2020}) \bibinfo{pages}{1546--1558}.
\bibitem[{Bajaj and Hasan(2019)}]{bajaj2019unipopt}
\bibinfo{author}{I.~Bajaj}, \bibinfo{author}{M.~F. Hasan},
\newblock \bibinfo{title}{Unipopt: Univariate projection-based optimization
  without derivatives},
\newblock \bibinfo{journal}{Computers \& Chemical Engineering}
  \bibinfo{volume}{127} (\bibinfo{year}{2019}) \bibinfo{pages}{71--87}.
\bibitem[{Eriksson et~al.(2019)Eriksson, Pearce, Gardner, Turner, and
  Poloczek}]{eriksson2019scalable}
\bibinfo{author}{D.~Eriksson}, \bibinfo{author}{M.~Pearce},
  \bibinfo{author}{J.~Gardner}, \bibinfo{author}{R.~D. Turner},
  \bibinfo{author}{M.~Poloczek},
\newblock \bibinfo{title}{Scalable global optimization via local {B}ayesian
  optimization},
\newblock \bibinfo{journal}{Advances in Neural Information Processing Systems}
  \bibinfo{volume}{32} (\bibinfo{year}{2019}).
\bibitem[{Deshwal et~al.(2021)Deshwal, Belakaria, and
  Doppa}]{deshwal2021bayesian}
\bibinfo{author}{A.~Deshwal}, \bibinfo{author}{S.~Belakaria},
  \bibinfo{author}{J.~R. Doppa},
\newblock \bibinfo{title}{Bayesian optimization over hybrid spaces},
\newblock in: \bibinfo{booktitle}{International Conference on Machine
  Learning}, \bibinfo{organization}{PMLR}, \bibinfo{year}{2021}, pp.
  \bibinfo{pages}{2632--2643}.
\bibitem[{Thebelt et~al.(2022)Thebelt, Tsay, Lee, Sudermann-Merx, Walz, Shafei,
  and Misener}]{thebelt2022tree}
\bibinfo{author}{A.~Thebelt}, \bibinfo{author}{C.~Tsay},
  \bibinfo{author}{R.~Lee}, \bibinfo{author}{N.~Sudermann-Merx},
  \bibinfo{author}{D.~Walz}, \bibinfo{author}{B.~Shafei},
  \bibinfo{author}{R.~Misener},
\newblock \bibinfo{title}{Tree ensemble kernels for {Bayesian} optimization
  with known constraints over mixed-feature spaces},
\newblock \bibinfo{journal}{Advances in Neural Information Processing Systems}
  \bibinfo{volume}{35} (\bibinfo{year}{2022}) \bibinfo{pages}{37401--37415}.
\bibitem[{Ru et~al.(2020)Ru, Alvi, Nguyen, Osborne, and
  Roberts}]{ru2020bayesian}
\bibinfo{author}{B.~Ru}, \bibinfo{author}{A.~Alvi},
  \bibinfo{author}{V.~Nguyen}, \bibinfo{author}{M.~A. Osborne},
  \bibinfo{author}{S.~Roberts},
\newblock \bibinfo{title}{Bayesian optimisation over multiple continuous and
  categorical inputs},
\newblock in: \bibinfo{booktitle}{International Conference on Machine
  Learning}, \bibinfo{organization}{PMLR}, \bibinfo{year}{2020}, pp.
  \bibinfo{pages}{8276--8285}.
\bibitem[{Daxberger et~al.(2021)Daxberger, Makarova, Turchetta, and
  Krause}]{daxberger2021mixed}
\bibinfo{author}{E.~Daxberger}, \bibinfo{author}{A.~Makarova},
  \bibinfo{author}{M.~Turchetta}, \bibinfo{author}{A.~Krause},
\newblock \bibinfo{title}{Mixed-variable {Bayesian} optimization},
\newblock in: \bibinfo{booktitle}{Proceedings of the Twenty-Ninth International
  Conference on International Joint Conferences on Artificial Intelligence},
  \bibinfo{year}{2021}, pp. \bibinfo{pages}{2633--2639}.
\bibitem[{Thebelt et~al.(2022)Thebelt, Tsay, Lee, Sudermann-Merx, Walz,
  Tranter, and Misener}]{thebelt2022multi}
\bibinfo{author}{A.~Thebelt}, \bibinfo{author}{C.~Tsay}, \bibinfo{author}{R.~M.
  Lee}, \bibinfo{author}{N.~Sudermann-Merx}, \bibinfo{author}{D.~Walz},
  \bibinfo{author}{T.~Tranter}, \bibinfo{author}{R.~Misener},
\newblock \bibinfo{title}{Multi-objective constrained optimization for energy
  applications via tree ensembles},
\newblock \bibinfo{journal}{Applied Energy} \bibinfo{volume}{306}
  (\bibinfo{year}{2022}) \bibinfo{pages}{118061}.
\bibitem[{Dreczkowski et~al.(2023)Dreczkowski, Grosnit, and
  Ammar}]{dreczkowski2023framework}
\bibinfo{author}{K.~Dreczkowski}, \bibinfo{author}{A.~Grosnit},
  \bibinfo{author}{H.~B. Ammar},
\newblock \bibinfo{title}{Framework and benchmarks for combinatorial and
  mixed-variable {B}ayesian optimization},
\newblock \bibinfo{journal}{arXiv preprint arXiv:2306.09803}
  (\bibinfo{year}{2023}).
\bibitem[{Rangaiah et~al.(2020)Rangaiah, Feng, and Hoadley}]{rangaiah2020multi}
\bibinfo{author}{G.~P. Rangaiah}, \bibinfo{author}{Z.~Feng},
  \bibinfo{author}{A.~F. Hoadley},
\newblock \bibinfo{title}{Multi-objective optimization applications in chemical
  process engineering: Tutorial and review},
\newblock \bibinfo{journal}{Processes} \bibinfo{volume}{8}
  (\bibinfo{year}{2020}) \bibinfo{pages}{508}.
\bibitem[{Folch et~al.(2023)Folch, Odgers, Zhang, Lee, Shafei, Walz, Tsay,
  van~der Wilk, and Misener}]{folch2023practical}
\bibinfo{author}{J.~P. Folch}, \bibinfo{author}{J.~Odgers},
  \bibinfo{author}{S.~Zhang}, \bibinfo{author}{R.~M. Lee},
  \bibinfo{author}{B.~Shafei}, \bibinfo{author}{D.~Walz},
  \bibinfo{author}{C.~Tsay}, \bibinfo{author}{M.~van~der Wilk},
  \bibinfo{author}{R.~Misener},
\newblock \bibinfo{title}{Practical path-based {B}ayesian optimization},
\newblock \bibinfo{journal}{arXiv preprint arXiv:2312.00622}
  (\bibinfo{year}{2023}).
\bibitem[{Beykal et~al.(2018)Beykal, Boukouvala, Floudas, and
  Pistikopoulos}]{beykal2018optimal}
\bibinfo{author}{B.~Beykal}, \bibinfo{author}{F.~Boukouvala},
  \bibinfo{author}{C.~A. Floudas}, \bibinfo{author}{E.~N. Pistikopoulos},
\newblock \bibinfo{title}{Optimal design of energy systems using constrained
  grey-box multi-objective optimization},
\newblock \bibinfo{journal}{Computers \& Chemical Engineering}
  \bibinfo{volume}{116} (\bibinfo{year}{2018}) \bibinfo{pages}{488--502}.
\bibitem[{Fromer et~al.(2023)Fromer, Graff, and Coley}]{fromer2023pareto}
\bibinfo{author}{J.~C. Fromer}, \bibinfo{author}{D.~E. Graff},
  \bibinfo{author}{C.~W. Coley},
\newblock \bibinfo{title}{Pareto optimization to accelerate multi-objective
  virtual screening},
\newblock \bibinfo{journal}{arXiv preprint arXiv:2310.10598}
  (\bibinfo{year}{2023}).
\bibitem[{Paulson and Lu(2022)}]{paulson2022cobalt}
\bibinfo{author}{J.~A. Paulson}, \bibinfo{author}{C.~Lu},
\newblock \bibinfo{title}{{COBALT}: {COnstrained Bayesian optimizAtion of
  computationaLly expensive grey-box models exploiting derivaTive
  information}},
\newblock \bibinfo{journal}{Computers \& Chemical Engineering}
  \bibinfo{volume}{160} (\bibinfo{year}{2022}) \bibinfo{pages}{107700}.
\bibitem[{Astudillo and Frazier(2021)}]{astudillo2021bayesian}
\bibinfo{author}{R.~Astudillo}, \bibinfo{author}{P.~Frazier},
\newblock \bibinfo{title}{Bayesian optimization of function networks},
\newblock \bibinfo{journal}{Advances in Neural Information Processing Systems}
  \bibinfo{volume}{34} (\bibinfo{year}{2021}) \bibinfo{pages}{14463--14475}.
\bibitem[{Gonz{\'a}lez and Zavala(2023)}]{gonzalez2023bois}
\bibinfo{author}{L.~D. Gonz{\'a}lez}, \bibinfo{author}{V.~M. Zavala},
\newblock \bibinfo{title}{{BOIS: B}ayesian optimization of interconnected
  systems},
\newblock \bibinfo{journal}{arXiv preprint arXiv:2311.11254}
  (\bibinfo{year}{2023}).
\bibitem[{Berkenkamp et~al.(2023)Berkenkamp, Krause, and
  Schoellig}]{berkenkamp2023bayesian}
\bibinfo{author}{F.~Berkenkamp}, \bibinfo{author}{A.~Krause},
  \bibinfo{author}{A.~P. Schoellig},
\newblock \bibinfo{title}{Bayesian optimization with safety constraints: Safe
  and automatic parameter tuning in robotics},
\newblock \bibinfo{journal}{Machine Learning} \bibinfo{volume}{112}
  (\bibinfo{year}{2023}) \bibinfo{pages}{3713--3747}.
\bibitem[{Krishnamoorthy and Doyle(2022)}]{krishnamoorthy2022safe}
\bibinfo{author}{D.~Krishnamoorthy}, \bibinfo{author}{F.~J. Doyle},
\newblock \bibinfo{title}{Safe {B}ayesian optimization using interior-point
  methods—applied to personalized insulin dose guidance},
\newblock \bibinfo{journal}{IEEE Control Systems Letters} \bibinfo{volume}{6}
  (\bibinfo{year}{2022}) \bibinfo{pages}{2834--2839}.
\bibitem[{Chan et~al.(2023)Chan, Paulson, and Mesbah}]{chanCDCsafeBO}
\bibinfo{author}{K.~J. Chan}, \bibinfo{author}{J.~A. Paulson},
  \bibinfo{author}{A.~Mesbah},
\newblock \bibinfo{title}{Safe explorative {B}ayesian optimization - towards
  personalized treatments in plasma medicine},
\newblock in: \bibinfo{booktitle}{Proceedings of Conference on Decision and
  Control (accepted)}, \bibinfo{organization}{IEEE}, \bibinfo{year}{2023}.
\bibitem[{Picheny et~al.(2016)Picheny, Gramacy, Wild, and
  Le~Digabel}]{picheny2016bayesian}
\bibinfo{author}{V.~Picheny}, \bibinfo{author}{R.~B. Gramacy},
  \bibinfo{author}{S.~Wild}, \bibinfo{author}{S.~Le~Digabel},
\newblock \bibinfo{title}{Bayesian optimization under mixed constraints with a
  slack-variable augmented {L}agrangian},
\newblock \bibinfo{journal}{Advances in Neural Information Processing Systems}
  \bibinfo{volume}{29} (\bibinfo{year}{2016}).
\bibitem[{Gardner et~al.(2014)Gardner, Kusner, Xu, Weinberger, and
  Cunningham}]{gardner2014bayesian}
\bibinfo{author}{J.~R. Gardner}, \bibinfo{author}{M.~J. Kusner},
  \bibinfo{author}{Z.~E. Xu}, \bibinfo{author}{K.~Q. Weinberger},
  \bibinfo{author}{J.~P. Cunningham},
\newblock \bibinfo{title}{Bayesian optimization with inequality constraints.},
\newblock in: \bibinfo{booktitle}{ICML}, volume \bibinfo{volume}{2014},
  \bibinfo{year}{2014}, pp. \bibinfo{pages}{937--945}.
\bibitem[{Lu and Paulson(2022)}]{lu2022no}
\bibinfo{author}{C.~Lu}, \bibinfo{author}{J.~A. Paulson},
\newblock \bibinfo{title}{No-regret {B}ayesian optimization with unknown
  equality and inequality constraints using exact penalty functions},
\newblock \bibinfo{journal}{IFAC-PapersOnLine} \bibinfo{volume}{55}
  (\bibinfo{year}{2022}) \bibinfo{pages}{895--902}.
\bibitem[{Lu and Paulson(2023)}]{LU2023103085}
\bibinfo{author}{C.~Lu}, \bibinfo{author}{J.~A. Paulson},
\newblock \bibinfo{title}{No-regret constrained {B}ayesian optimization of
  noisy and expensive hybrid models using differentiable quantile function
  approximations},
\newblock \bibinfo{journal}{Journal of Process Control} \bibinfo{volume}{131}
  (\bibinfo{year}{2023}) \bibinfo{pages}{103085}.
\bibitem[{Paulson et~al.(2022)Paulson, Sorouifar, and
  Chakrabarty}]{paulson2022efficient}
\bibinfo{author}{J.~A. Paulson}, \bibinfo{author}{F.~Sorouifar},
  \bibinfo{author}{A.~Chakrabarty},
\newblock \bibinfo{title}{Efficient multi-step lookahead {B}ayesian
  optimization with local search constraints},
\newblock in: \bibinfo{booktitle}{2022 IEEE 61st Conference on Decision and
  Control (CDC)}, \bibinfo{year}{2022}, pp. \bibinfo{pages}{123--129}.
\bibitem[{Paulson et~al.(2023)Paulson, Sorouifar, Laughman, and
  Chakrabarty}]{paulson2023lsr}
\bibinfo{author}{J.~A. Paulson}, \bibinfo{author}{F.~Sorouifar},
  \bibinfo{author}{C.~R. Laughman}, \bibinfo{author}{A.~Chakrabarty},
\newblock \bibinfo{title}{{LSR-BO}: Local search region constrained {B}ayesian
  optimization for performance optimization of vapor compression systems},
\newblock in: \bibinfo{booktitle}{Proceedings of the American Control
  Conference}, \bibinfo{organization}{IEEE}, \bibinfo{year}{2023}, pp.
  \bibinfo{pages}{576--582}.
\bibitem[{Paulson et~al.(2024)Paulson, Sorourifar, Laughman, and
  Chakrabarty}]{paulson2024self}
\bibinfo{author}{J.~A. Paulson}, \bibinfo{author}{F.~Sorourifar},
  \bibinfo{author}{C.~R. Laughman}, \bibinfo{author}{A.~Chakrabarty},
\newblock \bibinfo{title}{Self-optimizing vapor compression cycles online with
  {B}ayesian optimization under local search region constraints},
\newblock \bibinfo{journal}{Journal of Dynamic Systems, Measurement, and
  Control} \bibinfo{volume}{146} (\bibinfo{year}{2024}).
\bibitem[{Folch et~al.(2022)Folch, Zhang, Lee, Shafei, Walz, Tsay, van~der
  Wilk, and Misener}]{folch2022snake}
\bibinfo{author}{J.~P. Folch}, \bibinfo{author}{S.~Zhang},
  \bibinfo{author}{R.~Lee}, \bibinfo{author}{B.~Shafei},
  \bibinfo{author}{D.~Walz}, \bibinfo{author}{C.~Tsay},
  \bibinfo{author}{M.~van~der Wilk}, \bibinfo{author}{R.~Misener},
\newblock \bibinfo{title}{{SnAKe}: {B}ayesian optimization with pathwise
  exploration},
\newblock \bibinfo{journal}{Advances in Neural Information Processing Systems}
  \bibinfo{volume}{35} (\bibinfo{year}{2022}) \bibinfo{pages}{35226--35239}.
\bibitem[{Yang et~al.(2023)Yang, Aitchison, and Moss}]{yang2023mongoose}
\bibinfo{author}{A.~X. Yang}, \bibinfo{author}{L.~Aitchison},
  \bibinfo{author}{H.~B. Moss},
\newblock \bibinfo{title}{{MONGOOSE}: Path-wise smooth {B}ayesian optimisation
  via meta-learning},
\newblock \bibinfo{journal}{arXiv preprint arXiv:2302.11533}
  (\bibinfo{year}{2023}).
\bibitem[{Ramesh et~al.(2022)Ramesh, Sessa, Krause, and
  Bogunovic}]{ramesh2022movement}
\bibinfo{author}{S.~S. Ramesh}, \bibinfo{author}{P.~G. Sessa},
  \bibinfo{author}{A.~Krause}, \bibinfo{author}{I.~Bogunovic},
\newblock \bibinfo{title}{Movement penalized {B}ayesian optimization with
  application to wind energy systems},
\newblock \bibinfo{journal}{Advances in Neural Information Processing Systems}
  \bibinfo{volume}{35} (\bibinfo{year}{2022}) \bibinfo{pages}{27036--27048}.
\bibitem[{Foumani et~al.(2023)Foumani, Shishehbor, Yousefpour, and
  Bostanabad}]{foumani2023multi}
\bibinfo{author}{Z.~Z. Foumani}, \bibinfo{author}{M.~Shishehbor},
  \bibinfo{author}{A.~Yousefpour}, \bibinfo{author}{R.~Bostanabad},
\newblock \bibinfo{title}{Multi-fidelity cost-aware {Bayesian} optimization},
\newblock \bibinfo{journal}{Computer Methods in Applied Mechanics and
  Engineering} \bibinfo{volume}{407} (\bibinfo{year}{2023})
  \bibinfo{pages}{115937}.
\bibitem[{Folch et~al.(2023)Folch, Lee, Shafei, Walz, Tsay, van~der Wilk, and
  Misener}]{folch2023combining}
\bibinfo{author}{J.~P. Folch}, \bibinfo{author}{R.~M. Lee},
  \bibinfo{author}{B.~Shafei}, \bibinfo{author}{D.~Walz},
  \bibinfo{author}{C.~Tsay}, \bibinfo{author}{M.~van~der Wilk},
  \bibinfo{author}{R.~Misener},
\newblock \bibinfo{title}{Combining multi-fidelity modelling and asynchronous
  batch {B}ayesian optimization},
\newblock \bibinfo{journal}{Computers \& Chemical Engineering}
  \bibinfo{volume}{172} (\bibinfo{year}{2023}) \bibinfo{pages}{108194}.
\bibitem[{Zhou and Zhong(2023)}]{zhou2023comprehensive}
\bibinfo{author}{R.~Zhou}, \bibinfo{author}{J.~Zhong},
\newblock \bibinfo{title}{A comprehensive pragmatic investigation of batched
  acquisition functions in {Bayesian} optimization},
\newblock in: \bibinfo{booktitle}{Proceedings of the Companion Conference on
  Genetic and Evolutionary Computation}, \bibinfo{year}{2023}, pp.
  \bibinfo{pages}{831--834}.
\bibitem[{Hern{\'a}ndez-Lobato et~al.(2017)Hern{\'a}ndez-Lobato, Requeima,
  Pyzer-Knapp, and Aspuru-Guzik}]{hernandez2017parallel}
\bibinfo{author}{J.~M. Hern{\'a}ndez-Lobato}, \bibinfo{author}{J.~Requeima},
  \bibinfo{author}{E.~O. Pyzer-Knapp}, \bibinfo{author}{A.~Aspuru-Guzik},
\newblock \bibinfo{title}{Parallel and distributed thompson sampling for
  large-scale accelerated exploration of chemical space},
\newblock in: \bibinfo{booktitle}{International Conference on Machine
  Learning}, \bibinfo{organization}{PMLR}, \bibinfo{year}{2017}, pp.
  \bibinfo{pages}{1470--1479}.
\bibitem[{Adachi et~al.(2023{\natexlab{a}})Adachi, Hayakawa, Hamid,
  J{\o}rgensen, Oberhauser, and Osborne}]{adachi2023sober}
\bibinfo{author}{M.~Adachi}, \bibinfo{author}{S.~Hayakawa},
  \bibinfo{author}{S.~Hamid}, \bibinfo{author}{M.~J{\o}rgensen},
  \bibinfo{author}{H.~Oberhauser}, \bibinfo{author}{M.~A. Osborne},
\newblock \bibinfo{title}{{SOBER:} scalable batch {Bayesian} optimization and
  quadrature using recombination constraints},
\newblock \bibinfo{journal}{arXiv preprint arXiv:2301.11832}
  (\bibinfo{year}{2023}{\natexlab{a}}).
\bibitem[{Adachi et~al.(2023{\natexlab{b}})Adachi, Hayakawa, Wan, J{\o}rgensen,
  Oberhauser, and Osborne}]{adachi2023domain}
\bibinfo{author}{M.~Adachi}, \bibinfo{author}{S.~Hayakawa},
  \bibinfo{author}{X.~Wan}, \bibinfo{author}{M.~J{\o}rgensen},
  \bibinfo{author}{H.~Oberhauser}, \bibinfo{author}{M.~A. Osborne},
\newblock \bibinfo{title}{Domain-agnostic batch {B}ayesian optimization with
  diverse constraints via {B}ayesian quadrature},
\newblock \bibinfo{journal}{arXiv preprint arXiv:2306.05843}
  (\bibinfo{year}{2023}{\natexlab{b}}).
\bibitem[{Liang et~al.(2023)Liang, Hu, Han, Chen, and Yuan}]{liang2023capbo}
\bibinfo{author}{R.~Liang}, \bibinfo{author}{H.~Hu}, \bibinfo{author}{Y.~Han},
  \bibinfo{author}{B.~Chen}, \bibinfo{author}{Z.~Yuan},
\newblock \bibinfo{title}{{CAPBO}: A cost-aware parallelized {Bayesian}
  optimization method for chemical reaction optimization},
\newblock \bibinfo{journal}{AIChE Journal}  (\bibinfo{year}{2023})
  \bibinfo{pages}{e18316}.

\end{thebibliography}

\end{document}